\documentclass{article}

\makeatletter
\def\input@path{{styles/}}
\makeatother
\usepackage[preprint]{colm2026_conference}
\usepackage{fontspec}
\usepackage{xeCJK}
\setCJKsansfont[BoldFont=FandolHei-Bold.otf]{FandolHei-Regular.otf}
\setCJKmonofont{FandolFang-Regular.otf}

\normalfont
\usepackage[protrusion=false]{microtype}
\usepackage{graphicx}
\usepackage{xcolor}
\usepackage{amsmath}
\usepackage{amssymb}
\usepackage{booktabs}
\usepackage{array}
\usepackage{tabularx}
\usepackage{colortbl}
\usepackage{float}
\usepackage{placeins}
\usepackage{wrapfig}
\usepackage{tikz}
\usetikzlibrary{arrows.meta,positioning,fit,backgrounds,calc}
\usepackage{tcolorbox}
\usepackage{hyperref}
\usepackage{url}
\definecolor{abyss}{HTML}{121D36}
\definecolor{polarnight}{HTML}{1A2947}
\definecolor{nebula}{HTML}{2B3F66}
\definecolor{steeltrail}{HTML}{6D87BD}
\definecolor{skytrail}{HTML}{8FA8D8}
\definecolor{starlight}{HTML}{DFE7F5}
\definecolor{warmstar}{HTML}{E8D9C4}
\definecolor{allsparkwordmark}{HTML}{16233F}
\definecolor{allsparkspark}{HTML}{4A659C}
\definecolor{electricblue}{HTML}{3866FF}
\definecolor{covercream}{HTML}{EEF3FA}
\definecolor{coveraccent}{HTML}{3866FF}
\colorlet{pevekpurple}{skytrail}
\colorlet{bargray}{steeltrail}
\colorlet{barlgray}{starlight}

\definecolor{jrcolor}{RGB}{225, 100, 150}

\newfontfamily\outfit[
  Path=assets/fonts/,
  UprightFont=Outfit-Regular.ttf,
  BoldFont=Outfit-SemiBold.ttf
]{Outfit}

\newif\ifdraftnotes
\draftnotesfalse

\hypersetup{
  colorlinks=true,
  linkcolor=electricblue,
  citecolor=electricblue,
  urlcolor=coveraccent,
  filecolor=electricblue
}
\setcitestyle{numbers,square,comma,sort&compress}

\newcommand{\reportname}{AdaptEvo}
\newcommand{\reporttitleremainder}{Learning under Imperfect Supervision through Confidence-Adaptive Training with Knowledge Evolution}

\newcommand{\reporttitle}{\reportname: \reporttitleremainder}
\newcommand{\reportauthors}{AllSpark Team}
\newcommand{\reportauthorblock}{%
  \begingroup
  \centering
  {\fontsize{12}{15}\selectfont\bfseries \reportauthors\par}
  \endgroup
}
\newcommand{\reportdate}{September 30, 2026}
\title{\reporttitle}
\author{\reportauthors}
\hypersetup{pdftitle={\reporttitle},pdfauthor={\reportauthors},pdfsubject={Adaptive agent learning under imperfect supervision}}

\begin{document}

\fancyhead{}
\renewcommand{\headrulewidth}{0pt}
\fancyfoot[C]{\thepage}
\raggedbottom
\color{abyss}
\thispagestyle{empty}

\vspace*{-0.44in}
\begin{tcolorbox}[
  width=\linewidth,
  colback=covercream,
  colframe=covercream,
  boxrule=0pt,
  arc=14pt,
  outer arc=14pt,
  boxsep=0pt,
  left=20pt,
  right=20pt,
  top=13pt,
  bottom=11pt
]
  {\outfit\fontsize{19.5}{23}\selectfont\bfseries\centering
    \textcolor{coveraccent}{\reportname}\textcolor{black}{: \reporttitleremainder}\par}
  \vspace{0.7em}
  \reportauthorblock

  \vspace{0.75em}
  \begingroup
  \normalfont
  \setlength{\parindent}{0pt}
  \setlength{\parskip}{0pt}

Rule-governed contextual decision tasks require models to apply specified
rules to case-specific context and evidence. Written rules can leave gaps
in decision guidance and process evaluation, while reference judgments
vary in their support from the rules and evidence.
To address these challenges, we introduce \textbf{AdaptEvo}, a framework
for learning under imperfect supervision that couples confidence-adaptive
policy optimization with evolving decision knowledge and evaluation
rubrics. Its Training module uses Confidence-Adaptive GRPO (CA-GRPO) to balance
outcome and process rewards according to reference confidence. Its Evolution module
synthesizes reusable decision knowledge from recurring failures across
training cases and refines process rubrics to detect overlooked errors. To support empirical evaluation, 
we construct an industrial multimodal content moderation
dataset comprising a training set and In-Period and Out-of-Period test
sets, with the latter collected under changed rules.
Using Qwen3.6-35B-A3B, AdaptEvo achieves 61.9\% exact-label accuracy and
72.2\% binary decision accuracy on In-Period, exceeding GRPO by 7.5 and
3.7 percentage points, respectively. On Out-of-Period, the policy trained
with CA-GRPO retains exact-label accuracy gains over the base model across
evaluated checkpoints without injected decision knowledge, while GRPO
declines with continued training. CA-GRPO also outperforms the tested fixed
reward mixtures on both Out-of-Period metrics.

  \par
  \endgroup

  \vspace{0.65em}
  \noindent
  \begin{minipage}[b]{0.63\linewidth}
    \outfit\fontsize{8.4}{10.2}\selectfont
    \textbf{Date:} \reportdate \\
    \textbf{Github: \url{https://github.com/AllSpark-Research/AdaptEvo}} 
  \end{minipage}%
  \hfill
  \begin{minipage}[b]{0.33\linewidth}
    \raggedleft
    \raisebox{-0.30em}{\includegraphics[height=16pt]{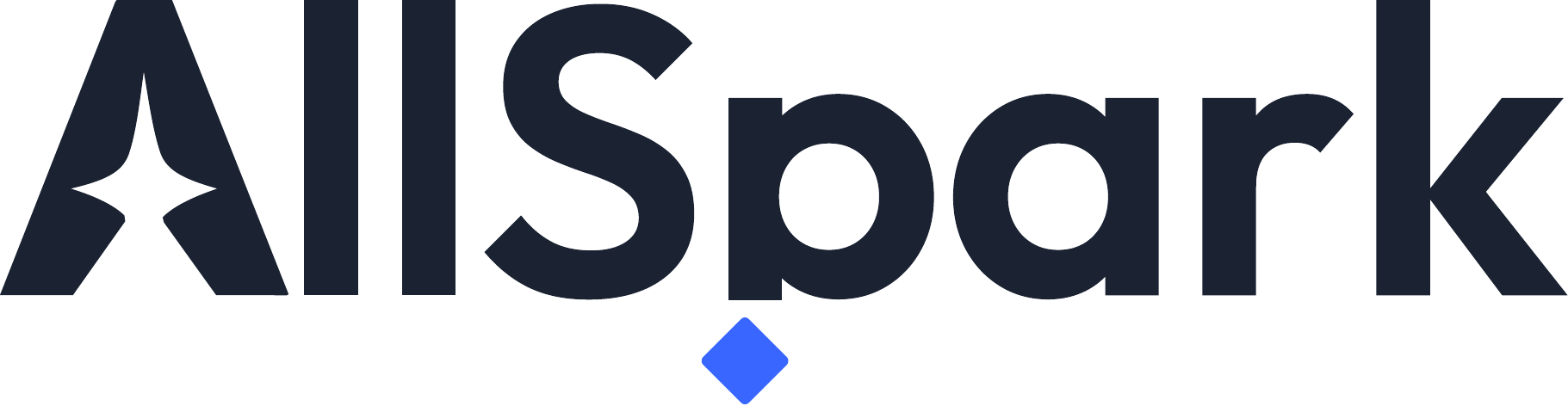}}%
  \end{minipage}
\end{tcolorbox}

\section{Introduction}
\label{sec:introduction}

\label{sec:introduction-motivation}

\begin{wrapfigure}[23]{r}{0.46\columnwidth}
    \vspace{\dimexpr-\intextsep+2pt\relax}
    \centering
    \includegraphics[width=\linewidth]{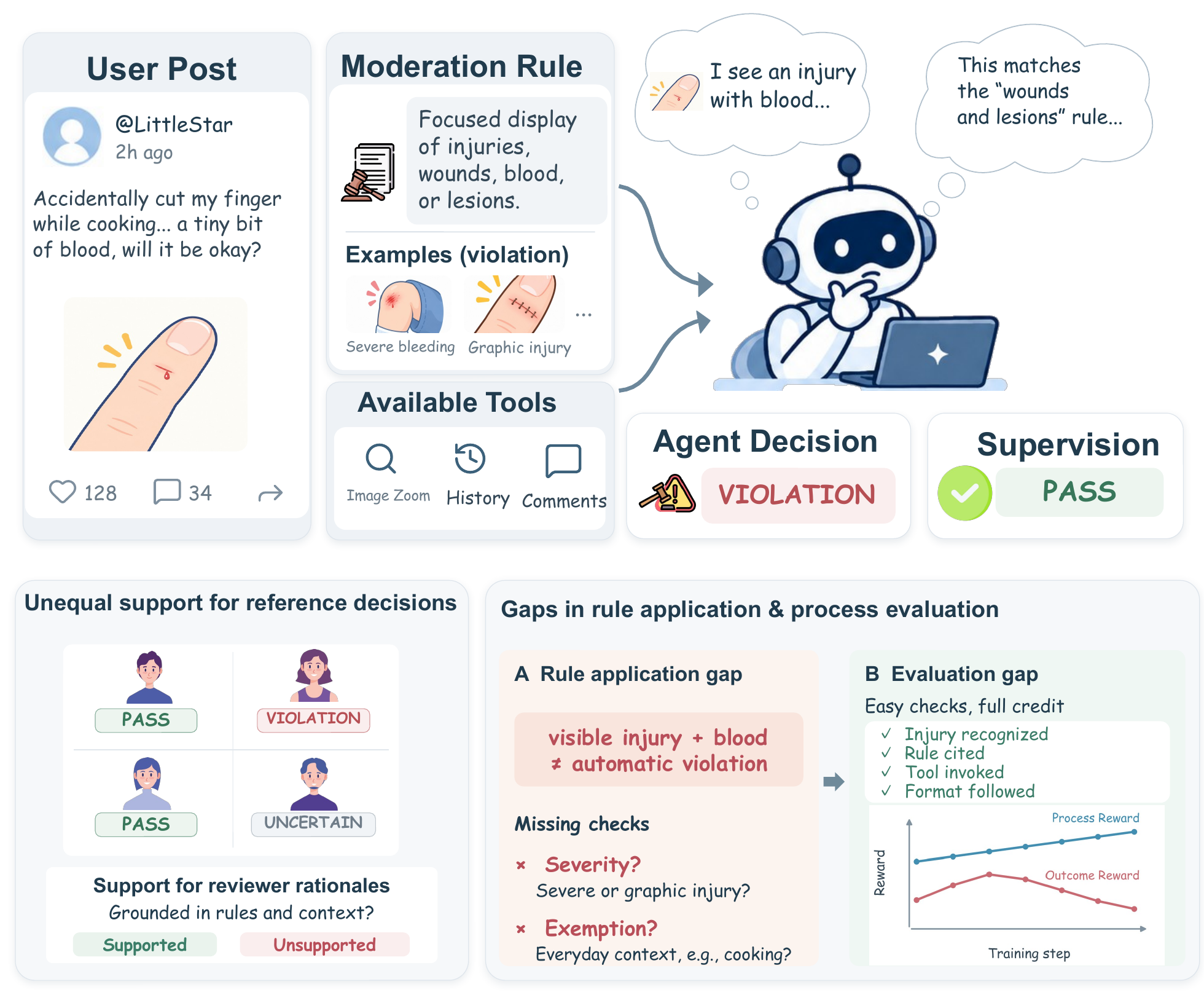}
    \fontsize{9}{10.5}\selectfont
    \caption{A content moderation example of a rule-governed
    contextual decision task, illustrating unequal support for reference
    decisions and gaps in rule application and process evaluation.
    The case and reward curves are schematic.}
    \label{fig:adaptevo-overview}
\end{wrapfigure}

Content moderation, legal judgment, and clinical guideline application require LLMs or agents to make decisions by applying written rules to specific contexts and available evidence~\citep{jacobs2021measurement,palla2025policy}. We refer to such tasks as rule-governed contextual decision tasks. These tasks primarily assess a model's ability to translate abstract rules into concrete decisions that are evidence-grounded and contextually appropriate.
RL-based training can use outcome feedback on final decisions and process feedback on the trajectories that produce them. Outcome rewards measure agreement with a reference label, whereas process rewards assess evidence acquisition, rule application, and decision derivation. The usefulness of these signals depends on the support for reference labels and the quality of process evaluation. 

Figure~\ref{fig:adaptevo-overview} illustrates two challenges in content moderation. On the one hand, written rules cannot exhaustively specify the evidence requirements, decision boundaries, and exceptions that arise across diverse contexts. This limitation not only leaves models without sufficient operational guidance but also makes it difficult for process evaluators with fixed criteria to detect errors in rule application. On the other hand, reference judgments vary in reliability: annotators may reach different conclusions because of differences in interpretation or implicit decision standards. Treating all reference labels as equally reliable outcome supervision may therefore cause models to learn decisions that are weakly supported by the rules and available evidence.
\label{sec:introduction-approach}

To address these challenges, we introduce \textbf{AdaptEvo}, a framework for learning under imperfect supervision in rule-governed contextual decision tasks. AdaptEvo alternates between two modules. In the \textbf{training module}, we propose Confidence-Adaptive GRPO (CA-GRPO), which adaptively balances outcome and process rewards according to reference confidence. 

CA-GRPO places greater emphasis on outcome feedback for strongly supported references and relies more on process feedback when reference judgments are uncertain. The \textbf{evolution module} freezes the updated policy. A stronger external model analyzes recurring failures across training cases to derive reusable decision conditions, evidence requirements, boundaries, and exceptions. Separately, rubric screening uses scores across cases and repeated judgments to select checks that reliably distinguish rollout quality. The evolved decision knowledge and evaluation rubric are then fixed and used to guide the next stage of training.

\label{sec:introduction-contributions}

To evaluate AdaptEvo, we construct a multimodal content moderation dataset
from real-world industrial workflows, with test sets from the training
period (In-Period) and a later period with updated rules (Out-of-Period).
Using Qwen3.6-35B-A3B, AdaptEvo achieves 61.9\% exact-label accuracy (ELA)
and 72.2\% binary decision accuracy (BDA) on In-Period, exceeding GRPO
by 7.5 and 3.7 percentage points, respectively.
On Out-of-Period, it achieves 59.5\% ELA and 69.5\% BDA, surpassing GRPO
by 7.6 and 4.3 points.
Further ablation studies and analyses support the benefits of evolved decision knowledge
and confidence-adaptive reward mixing.

Overall, our contributions are fourfold:
(1) \textbf{AdaptEvo}, a Training--Evolution framework for jointly adapting policy learning, decision guidance, and evaluation under imperfect supervision;
(2) \textbf{CA-GRPO}, a confidence-adaptive policy optimization method that balances outcome and process rewards according to reference confidence;
(3) training and test datasets from an industrial multimodal content moderation setting to support research on rule-governed contextual decision tasks;
and 
(4) empirical results demonstrating that AdaptEvo achieves state-of-the-art
performance on multimodal content moderation.

\section{Related Work}
\label{sec:related-work}
\paragraph{Imperfect supervision and decision criteria.}
\label{sec:related-supervision}
Annotator disagreement can reflect valid perspectives rather than noise
\citep{davani2022disagreements,pavlick2019disagreements}, while measurement and
policy research examines how abstract standards become operational judgments
\citep{jacobs2021measurement,palla2025policy}. AdaptEvo adjusts reward mixing
using support for a specified reference and learns scoped decision guidance.
It does not model the full distribution of perspectives or resolve standards
left undefined by the task.

\paragraph{Knowledge updates.}
\label{sec:related-agent-learning}
Self-improving systems update agent implementations or adaptation mechanisms \citep{robeyns2025sica,zhang2025dgm,zweiger2025seal}. Closer to our setting, AgentEvolver combines experience, RL, and reward attribution \citep{zhai2025agentevolver}; MemSkill alternates controller training with memory-skill revision \citep{zhang2026memskill}. AdaptEvo couples confidence-adaptive training with updates to decision knowledge and evaluation rubrics. Our two-stage evaluation covers one joint update, without establishing sustained recursive self-improvement.

\paragraph{Evaluation and feedback updates.}
\label{sec:related-evolving-feedback}
Reward overoptimization and judge biases motivate revisiting evaluation
signals as policies change
\citep{gao2022overoptimization,zheng2023judging}. Self-Rewarding Language Models
use model-generated preference labels for iterative preference optimization
\citep{yuan2024selfrewarding}. AdaptEvo instead revises textual process checks
and scoring anchors, with evaluator parameters held fixed.

\section{Preliminaries}
\label{sec:problem}

We use \textit{rule-governed contextual decision tasks} to denote tasks in which an agent applies explicitly provided rules or guidelines to case-specific context and evidence. Solving these tasks requires identifying relevant evidence, determining which rules apply, and checking the conditions and exceptions that affect the final decision. For example, in content moderation, observing a visible injury does not by itself establish a violation. The agent must assess its severity and presentation and determine whether any exemption applies under the governing rules. Such tasks therefore require contextual rule application beyond recognizing surface-level cues.

Given governing rules $R$ and a case context $x$, the agent must select
a label $\hat{y}\in\mathcal{Y}(x)$, where $\mathcal{Y}(x)$ denotes the
set of candidate labels for the case.
The predicted label should be supported by the available evidence and satisfy the applicable conditions and exceptions specified by $R$.

In our setting, we formulate this problem as an agentic task.
The initial context $x$ contains the available case information and potentially multimodal evidence.
A policy $\pi_\theta$ can interact with a tool set $\mathcal{T}$ to acquire
additional evidence before returning a label $\hat{y}$ and an explicit
rationale $j$.
The resulting observable trajectory is
\begin{equation}
    \tau = \bigl((a_1,o_1),\ldots,(a_K,o_K),j,\hat{y}\bigr),
    \label{eq:problem-trajectory}
\end{equation}
where $a_k$ is a request to a tool in $\mathcal{T}$, $o_k$ is its returned observation, and $K$ is the number of tool interactions.
Each action and the final decision are conditioned on the rules, initial
context, observations collected so far, and any injected decision knowledge.

For an annotated case, $y^\star$ denotes the reference label, optionally accompanied by a rationale or additional annotations.
The learning objective is to improve evidence-grounded decisions under $R$, with reference-label agreement serving as an imperfect supervision signal.

In this setting, learning faces two challenges arising from gaps in rule application and uncertainty in reference judgments.

\textbf{Unequal support for reference decisions.}
\label{sec:uneven-supervision}
Studies of subjective annotation and toxicity judgments show that
disagreement can reflect systematic interpretations and annotator beliefs
\citep{davani2022disagreements,sap2022attitudes}; these findings do not
make disagreement a correctness test. Under a specified rule, support must
also be assessed against the available evidence. In our moderation training
data, the final quality-control (QC) decision serves as the reference label
$y^\star$ (Appendix~\ref{sec:moderation-supervision}). Of $3{,}395$ cases, $709$
(\textbf{20.9\%}) have disagreement among the four human annotations.
Separately, Kimi-K3 assesses each reference label and its
written rationale against the applicable rules, post content, and cached
tool outputs. Across the same dataset, it rates $1{,}124$ cases
(\textbf{33.1\%}) as ambiguous or unsupported: $659$ (19.4\%) are ambiguous
and $465$ (13.7\%) are unsupported
(Appendix~\ref{app:supervision-diagnostics}). These are
model-assessed support judgments, not annotation-error estimates.
These partially overlapping case groups motivate adapting reliance on reference
labels while preserving useful process information.

\textbf{Gaps in rule application and process evaluation.}
\label{sec:implicit-standards}
\label{sec:process-reward-hacking}
Written rules may leave case-specific conditions and exceptions implicit
\citep{guitton2025opentexture,codish2005ambiguity,palla2025policy}.
For example, under a rule prohibiting severe graphic injury but exempting
medical education, an agent may flag a wound without checking severity
or the exemption. A rubric that rewards cue recognition and rule citation
may still score this decision highly.
Such failures motivate jointly refining decision knowledge and evaluation
rubrics so that agents apply the necessary checks and evaluators detect
their omission, with both grounded in the governing rules and evidence.

AdaptEvo addresses these challenges through \textbf{Training}, which adapts
outcome--process feedback to reference support, and \textbf{Evolution},
which revises decision knowledge and evaluation checks. The governing
rules remain fixed throughout the updates within a training run.

\section{Method}
\label{sec:method}

AdaptEvo maintains the state
\begin{equation}
    S_t = (\theta_t, E_t, Q_t),
    \label{eq:audit-state}
\end{equation}
where $t$ indexes training stages, $\theta_t$ denotes policy parameters,
$E_t$ is the decision knowledge library, and $Q_t$ is the process rubric.
CA-GRPO adapts policy learning to reference support; evolution updates
decision guidance and process feedback. A \emph{task scope} groups cases
sharing a candidate-label space and applicable rule context;
in our moderation setting, it corresponds to a moderation queue.

Each \textbf{training stage} (the inner loop) updates the policy with $E_t$ and
$Q_t$ fixed. At its boundary, an \textbf{evolution update} (the outer-loop
update) freezes the policy, synthesizes knowledge from recurring failures,
and screens rubric criteria using scores on saved training trajectories.
The selected versions define the next stage.
Knowledge guides the agent's application of governing rules $R$;
rubrics assess its behavior against those criteria. The applicable rule
versions, tools $\mathcal{T}$, and confidence definition remain fixed across
the training stages and evolution updates within a training run. Evaluation
may use different rule versions.

Our experiments cover two training stages and one joint knowledge--rubric
update. The full stage-transition protocol is presented in
Appendix~\ref{app:method-details}.

\begin{figure}[t]
\centering
\includegraphics[width=\linewidth]{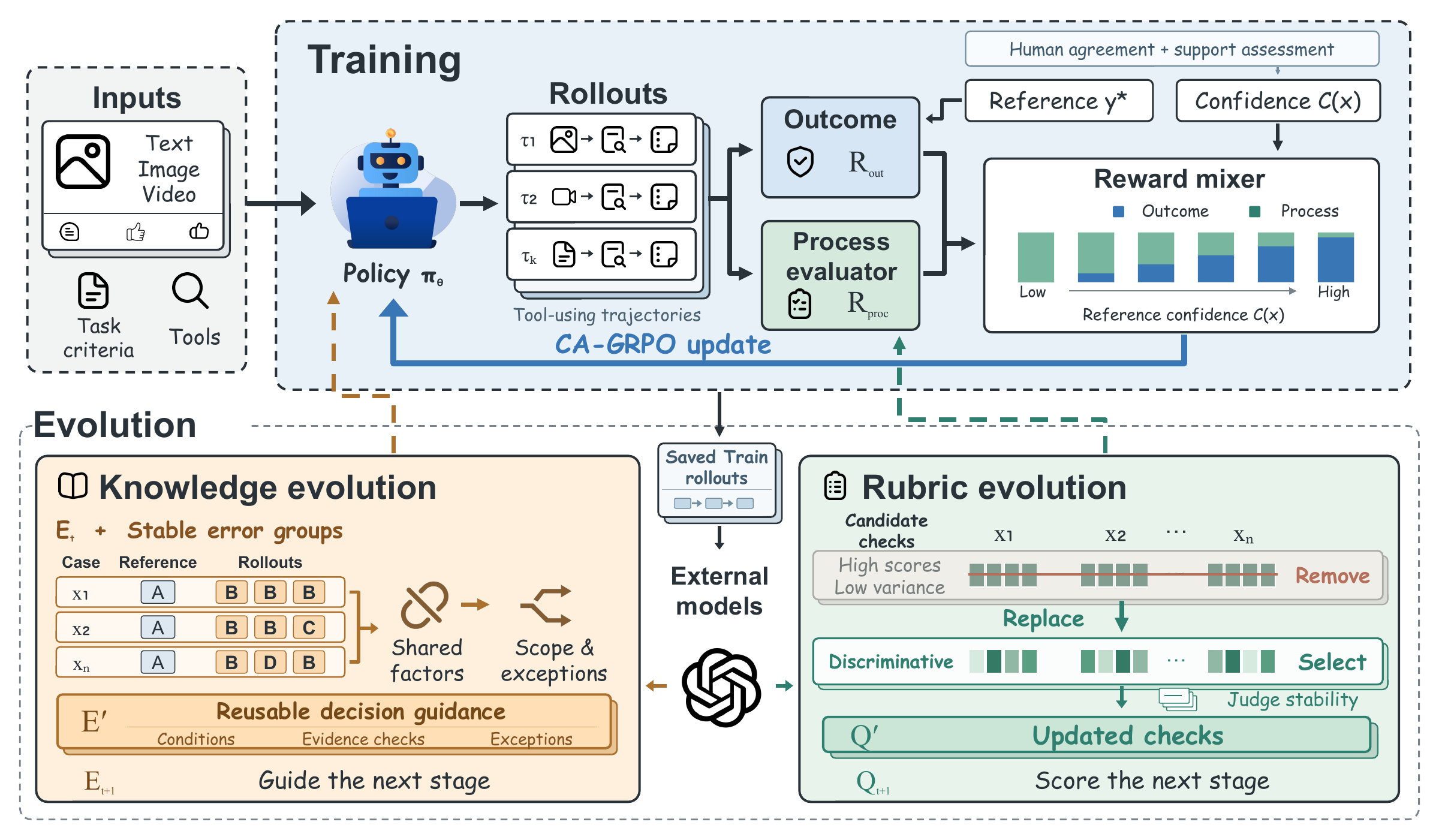}
\caption{AdaptEvo alternates training (top) and evolution (bottom).
Knowledge evolution turns recurring mismatches across cases (reference A,
predominantly B predictions) into conditional guidance. Rubric evolution
screens out saturated checks that lack within-case discrimination across
many cases, accounting for Judge noise when selecting alternatives. Each rubric
row is a check; each strip shows one case's rollout scores.
Labels and scores are schematic; task rules remain fixed.}
\label{fig:audit-rsi-framework}
\end{figure}

\subsection{Training: Confidence-Adaptive Policy Learning}
\label{sec:method-train}
\label{sec:confidence-adaptive-rl}

Confidence-Adaptive GRPO (CA-GRPO) adjusts the balance between reference
agreement and process feedback. Let $C(x)\in[0,1]$ measure support for reference
label $y^\star$ under the task criteria and available evidence. In our
implementation, it combines human annotation agreement with an external
assessment of support for the reference label and rationale.
It is distinct from task difficulty and need not be a calibrated correctness
probability. For a valid final output, the reward is
\begin{equation}
    R_t(\tau,x) = \alpha(C(x))R_{\mathrm{out}}(\tau,x)
    + [1-\alpha(C(x))]R_{\mathrm{proc}}(\tau;Q_t)
    - P_{\mathrm{protocol}}(\tau),
    \label{eq:adaptive-reward}
\end{equation}
where $R_{\mathrm{out}}$ measures reference agreement and
$P_{\mathrm{protocol}}$ penalizes tool and interaction-protocol errors.
Invalid final outputs receive zero reward. We use the fixed
mapping $\alpha(C)=0.9C$: adaptation varies across inputs, and process
feedback remains active even at maximal confidence.

The \emph{process evaluator} uses $Q_t$ to assess observable trajectories
against the applicable criteria and available evidence, excluding the raw
reference answer by default. It differs from the offline
\emph{reference-support assessor} used for confidence and data filtering.
We suppress input and rule conditioning in the notation:
\begin{equation}
    R_{\mathrm{proc}}(\tau;Q_t)
    = \sum_{d=1}^{D} w_d q_{d,t}(\tau),
    \qquad w_d\geq 0,\quad \sum_{d=1}^{D}w_d=1.
    \label{eq:process-reward}
\end{equation}
Here $D$ is the number of process dimensions, and $q_{d,t}(\tau)\in[0,1]$
is the score assigned to trajectory $\tau$ on dimension $d$ under $Q_t$.
During policy training, the evaluator assigns one score per dimension
using rubric checks and scoring anchors. Offline candidate-criterion
scores used for rubric screening are not separate training-reward terms.
The training dimensions and weights stay fixed across rubric revisions.
Appendix~\ref{sec:moderation-rewards} specifies the confidence mapping,
reward levels, output validation, and missing-score handling.

We optimize the mixed reward using standard GRPO~\citep{shao2024deepseekmath}.
Confidence changes the relative contributions of outcome and process
feedback within each rollout group, potentially altering trajectory rankings
when process scores distinguish rollouts.

\subsection{Evolution: Decision Knowledge and Evaluation Rubrics}
\label{sec:method-evolution}

\label{sec:failure-mining}
With the policy frozen, evolution uses saved training trajectories through
two complementary routes: cross-case mismatch analysis for knowledge and
multi-case score analysis for rubrics. The knowledge synthesizer and rubric
evaluator may be different external models.
\paragraph{Decision knowledge.}
\label{sec:experience-evolution}
Summarizing cases independently can turn reviewer-specific judgments or
thresholds into mutually inconsistent guidance. We first collect cases
within a task scope where repeated rollouts show similar disagreements
with reference labels, such as reference A with predominantly B predictions
(Figure~\ref{fig:audit-rsi-framework}). A stronger external model
$M_{\mathrm{strong}}$ jointly examines their rules and evidence;
diagnostic records $F_t$ distinguish decision defects from reference
conflicts or missing information. Repeated disagreement alone does not
establish that a reference is correct. A \emph{failure mechanism} is a
recurring, evidence-attributed reason for
a decision or process defect. $\mathcal{G}_t$ groups different training
inputs sharing a diagnosed mechanism, unlike GRPO's within-input rollout
groups. For each group, the model extracts shared evidence patterns and
failure factors, then contrasts context, evidence sufficiency, exceptions,
and neighboring-label boundaries. Only rule- and evidence-supported
distinctions enter the guidance; unresolved conflicts remain flagged for review.

The resulting guidance specifies triggering conditions, required evidence,
decision boundaries, and exceptions. Compatible findings from different
error groups are consolidated into one candidate for the task scope:
\begin{equation}
    E' = \mathcal{U}_E(E_t,\mathcal{G}_t;M_{\mathrm{strong}}).
    \label{eq:experience-update}
\end{equation}
For example, repeated attribution failures can yield guidance to verify
that a retrieved fact concerns the entity being judged, with explicit
conditions for transferring evidence. The injected text contains reusable
conditions and checks, with case identifiers and individual answers kept
outside it. Relevant knowledge enters the agent's initial context under a
fixed text budget, guiding the full trajectory. The knowledge-entry
structure appears in Appendix~\ref{app:experience-schema}.
\paragraph{Evaluation rubrics.}
\label{sec:rubric-evolution}
Rubric checks specify which observable behaviors to assess; \emph{scoring
anchors} specify which behaviors support particular score levels.
We screen a pool of candidate criteria $\mathcal{P}$ using saved
within-input rollout groups $\mathcal{B}_t$. A fixed evaluator scores the
criteria in batches and repeats judgments on the same trajectories.
Mean scores and full-score rates reveal saturation. We compute score
variance among rollouts within each input, then aggregate across cases
and queues to identify criteria that consistently offer little
discrimination. Comparison with matched repeated-judgment variance
accounts for evaluator noise; low variance on one case alone is insufficient.
Checks with high scores and low within-case variance across many cases are
candidates for replacement, subject to applicability and evidence review.
Alternatives with stable discrimination are re-evaluated on a separate
training panel before integration into
$Q'=\mathcal{U}_Q(Q_t,\mathcal{B}_t;\mathcal{P})$.
This revises checks and scoring anchors within the existing dimensions;
the five training dimensions and their weights remain fixed.
Appendix~\ref{app:rubric-screening}
details the screening and validation protocol.

Knowledge candidates require fresh agent rollouts because guidance can
change the full trajectory. Their assessment checks the targeted errors
and regressions on other training-side cases within the task scope;
rubric candidates reuse fixed trajectories. Selection yields $E_{t+1}$
and $Q_{t+1}$ from the candidates $E'$ and $Q'$ or retains either previous
version. Both remain fixed throughout the next training stage. Final test data
are excluded from diagnosis and update selection; implementation details
appear in Appendix~\ref{app:method-details}.

\section{Experiments}
\label{sec:experiments}
\subsection{Experimental Setup}
\label{sec:experiments-setup}

\paragraph{Task and data.}
\label{sec:experiments-questions}
We evaluate AdaptEvo in multimodal content moderation. Each case provides
a post (text and available images or videos), its moderation queue,
applicable rules, and access to tools such as comment retrieval and
similar-post search. The queue defines the review scope and candidate
labels: one pass label and a set of violation labels. The agent returns
one candidate label and a rationale grounded in the rules and evidence
(Appendix~\ref{app:moderation-task-instance} provides an example).

The training set contains 3,395 cases, each annotated by four human reviewers, including a quality-control (QC) reviewer. We use the QC reviewer’s label as the reference judgment, accompanied by a written rationale. Human annotations disagree on 20.9\% of the cases, highlighting the uncertainty in reference supervision. Reference confidence combines agreement among the human annotations with an offline assessment by Kimi-K3 of whether the rules and available evidence support the QC label and rationale. This assessment does not alter the reference label. Detailed annotation statistics are provided in Appendix~\ref{app:supervision-diagnostics}.

\paragraph{Models and comparisons.}
We evaluate Qwen, GLM, Kimi, and Gemini models on moderation label
prediction, then compare training methods using Qwen3.6-35B-A3B as the
shared backbone. Alongside the base model, we evaluate GRPO and CA-GRPO
trained on the full pool, and GRPO (High-Conf.) trained only on cases with
unanimous human labels and a QC label--rationale pair judged supported by
Kimi-K3. Both GRPO baselines use reference-based outcome rewards;
CA-GRPO uses confidence-adaptive outcome--process rewards with a fixed
rubric and no injected decision knowledge.
We also evaluate the same CA-GRPO policy with evolved knowledge only at
inference. AdaptEvo continues training from this policy with the same
knowledge and an updated process rubric.

\paragraph{Evaluation.}
Evaluation uses an \emph{In-Period} test set collected during the
training-data collection period and an \emph{Out-of-Period} test set
collected later under changed governing rules. We report
\emph{exact-label accuracy} (ELA), requiring an exact reference-label match,
and \emph{binary decision accuracy} (BDA), comparing pass/violation
decisions after merging all violation labels. Both metrics measure
reference agreement, not independently verified rule compliance;
reported differences are percentage points.
Appendix~\ref{app:experimental-details} provides annotation, reward,
metric, and training and evaluation details.

\subsection{Main Results}
\label{sec:results}

\label{sec:results-main}
\label{sec:results-reference-models}
Table~\ref{tab:main-in-period} reports results on the In-Period and
Out-of-Period benchmarks, with decision knowledge indicated separately.
We include reference models from different families and compare training
methods using Qwen3.6-35B-A3B as a shared backbone.

\begin{table}[t]
\normalsize
\setlength{\tabcolsep}{4pt}
\renewcommand{\arraystretch}{1.10}
\caption{Moderation results on the In-Period and Out-of-Period benchmarks
(\%, higher is better).
ELA: exact-label accuracy; BDA: binary decision accuracy.
Knowledge specifies injected decision knowledge. Bold marks the best score in each column.}
\label{tab:main-in-period}
\begin{center}
\begin{tabularx}{\linewidth}{@{}lc*{4}{>{\centering\arraybackslash}X}@{}}
\toprule
\textbf{Model / Method} & \textbf{Knowledge} & \multicolumn{2}{c}{\textbf{In-Period}} & \multicolumn{2}{c@{}}{\textbf{Out-of-Period}} \\
\cmidrule(lr){3-4}\cmidrule(l){5-6}
& & \textbf{ELA} & \textbf{BDA} & \textbf{ELA} & \textbf{BDA} \\
\midrule
Qwen3.5-4B & None & 46.8 & 60.8 & 50.8 & 64.5 \\
Qwen3.5-9B & None & 46.2 & 63.2 & 50.4 & 65.2 \\
Qwen3.5-27B & None & 55.1 & 68.5 & 53.8 & 66.0 \\
Qwen3.8-Flash-Next & None & 54.9 & 69.1 & 54.4 & 67.6 \\
GLM5.3-Flash & None & 56.3 & 69.6 & 55.6 & 67.5 \\
Kimi-K3 & None & 56.2 & 68.9 & 56.7 & 68.8 \\
Gemini3.8-Flash & None & 61.2 & 71.9 & \textbf{60.4} & 69.5 \\
\midrule
\multicolumn{6}{@{}l}{\textit{Shared backbone: Qwen3.6-35B-A3B}} \\
Qwen3.6-35B-A3B (base) & None & 53.9 & 66.1 & 53.3 & 65.9 \\
GRPO & None & 54.4 & 68.5 & 51.9 & 65.2 \\
GRPO (High-Conf.) & None & 53.3 & 68.8 & 52.9 & 65.5 \\
CA-GRPO & None & 55.4 & 68.1 & 55.2 & 67.3 \\
CA-GRPO$^{\dagger}$ & Evolved & 59.1 & 71.3 & 58.6 & \textbf{69.7} \\
\rowcolor{starlight!45}[0pt][0pt]
\textbf{AdaptEvo} & Evolved & \textbf{61.9} & \textbf{72.2} & 59.5 & 69.5 \\
\bottomrule
\end{tabularx}
\par\smallskip
\begin{minipage}{\linewidth}
\footnotesize
For each trained method, we report its 60-step checkpoint.
$^{\dagger}$The same CA-GRPO policy receives knowledge only at inference,
summarized by GPT-5.6-Sol.
AdaptEvo continues training from CA-GRPO with the same knowledge
and an updated rubric.
Run aliases and checkpoint conventions appear in Appendix~\ref{app:reproducibility}.
\end{minipage}
\end{center}
\end{table}

\paragraph{In-Period performance.}
Among the open-weight Qwen3.5 models, the 27B variant performs best,
reaching 55.1\% ELA and 68.5\% BDA.
AdaptEvo achieves \textbf{61.9\% ELA} and \textbf{72.2\% BDA}, the
highest In-Period results among all evaluated configurations, exceeding
the Qwen3.6-35B-A3B base model by 8.0/6.1 percentage points and GRPO
by 7.5/3.7 points. It also surpasses Gemini3.8-Flash, the strongest
reference model, by 0.7/0.3 points.
These gains reflect the complete configuration, including evolved
knowledge, rubric revision, and continued training.

\paragraph{Reward design and knowledge updates.}
On In-Period, GRPO (High-Conf.) trades 1.1 points of ELA for 0.3 points
of BDA relative to GRPO; CA-GRPO gains 1.0 ELA point but loses 0.4 BDA
points, motivating the fixed-mixture comparison in
Section~\ref{sec:analysis-confidence}.
Adding evolved knowledge at inference improves CA-GRPO by
3.7/3.2 ELA/BDA points. Continuing training with this knowledge and an
updated rubric yields AdaptEvo and adds 2.8/0.9 points.
These increments measure inference-time knowledge gains and the combined
effect of continued training and rubric revision, respectively.

\paragraph{Out-of-Period generalization.}
AdaptEvo achieves the highest Out-of-Period ELA among the evaluated
shared-backbone training methods at \textbf{59.5\%}, alongside 69.5\% BDA,
exceeding the base model by 6.2/3.6 percentage
points and GRPO by 7.6/4.3 points, respectively.
Without injected knowledge, CA-GRPO improves both metrics over the
base model (55.2/67.3 versus 53.3/65.9), whereas GRPO falls below the
base model on both (51.9/65.2). Adding evolved knowledge to CA-GRPO
further raises BDA to \textbf{69.7\%}, the highest among all evaluated
configurations. AdaptEvo also approaches Gemini3.8-Flash in ELA
(59.5\% versus 60.4\%) and matches its BDA (69.5\%).
Section~\ref{sec:analysis-generalization} examines these trends during
training: accuracy declines under GRPO despite higher training rewards,
whereas CA-GRPO retains ELA gains over the base model.

\section{Analysis}
\label{sec:analysis}
\subsection{Generalization to New Rules and Queues}
\label{sec:results-temporal}
\label{sec:analysis-generalization}

We track training rewards and checkpoint performance on Out-of-Period. All checkpoints are evaluated without injected decision
knowledge, and those trained with CA-GRPO precede knowledge and rubric
evolution.

\begin{figure}[t]
\centering
\includegraphics[width=\linewidth]{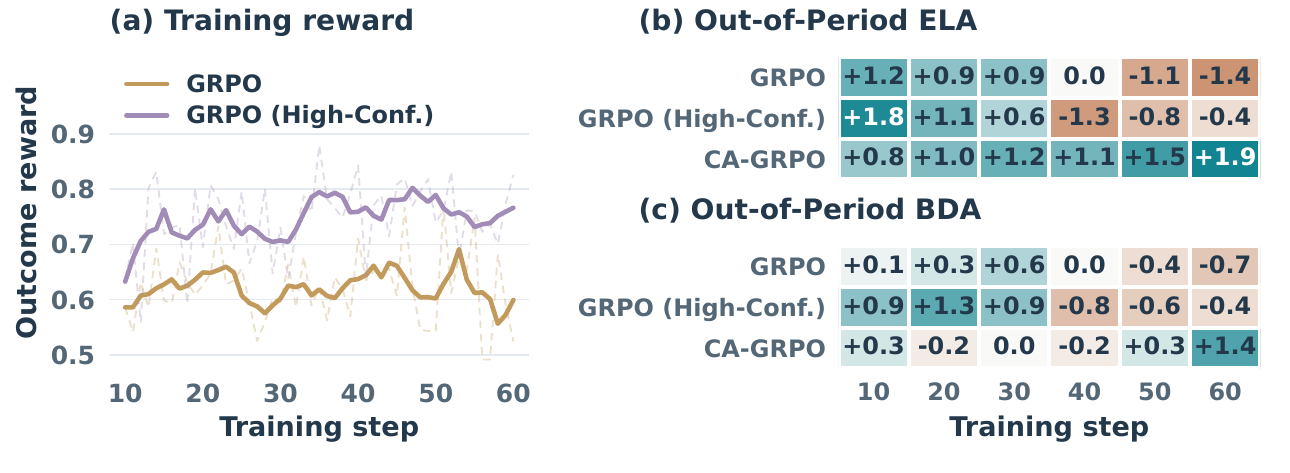}
\caption{Training reward and Out-of-Period generalization. (a) Outcome
rewards during GRPO and GRPO (High-Conf.) training: faint dashed lines show
raw values and solid lines show centered 5-step moving averages.
(b--c) Out-of-Period ELA and BDA gains over the base model
(53.3\%/65.9\%), in percentage points, without injected decision knowledge.
Heatmaps share a color scale: teal indicates gains and orange indicates
losses; each cell gives the exact difference.}
\label{fig:shifted-generalization}
\end{figure}

Figure~\ref{fig:shifted-generalization} shows that higher training reward
does not ensure better transfer. Despite substantial fluctuations, average
training rewards under GRPO and GRPO (High-Conf.) are higher in the final
training window than in the initial window. Out-of-Period checkpoint
accuracy nevertheless declines: from step 10 to 60, ELA/BDA fall by
2.6/0.8 percentage points under GRPO and 2.2/1.3 points under
GRPO (High-Conf.). The final checkpoints of both runs score below the base
model on both metrics. High-confidence filtering thus improves early
checkpoint performance but does not prevent subsequent generalization
degradation in this comparison.

The policy trained with CA-GRPO maintains an ELA gain over the base model
at every evaluated checkpoint. Its final checkpoint achieves the strongest
performance at 55.2\% ELA and 67.3\% BDA, exceeding the base model by
1.9/1.4 points and the final GRPO checkpoint by 3.3/2.1 points.
These results support stronger generalization as training proceeds.
The comparison concerns the complete CA-GRPO training
configuration; it does not isolate confidence adaptation from process
supervision. Appendix~\ref{app:shifted-evaluation} provides all checkpoint
values and reward-processing details.

\newpage
\subsection{Confidence-Adaptive Reward Mixing}
\label{sec:analysis-confidence}

\begin{wraptable}[14]{R}{0.48\textwidth}
\setlength{\tabcolsep}{3pt}
\renewcommand{\arraystretch}{1.08}
\caption{Fixed and adaptive reward mixing after 60 training steps.
Ratios are outcome:process; ELA and BDA are percentages.
Bold marks each column maximum.}
\label{tab:reward-mixing-main}
\begin{center}
\footnotesize
\begin{tabular}{@{}lrrrr@{}}
\toprule
& \multicolumn{2}{c}{In-Period} & \multicolumn{2}{c}{Out-of-Period} \\
\cmidrule(lr){2-3}\cmidrule(l){4-5}
Reward Mixing & ELA & BDA & ELA & BDA \\
\midrule
Fixed (0.8:0.2) & 54.3 & 67.6 & 53.6 & 66.1 \\
Fixed (0.5:0.5) & 54.2 & 67.9 & 54.0 & 66.2 \\
Fixed (0.2:0.8) & \textbf{55.6} & 67.3 & 54.8 & 65.8 \\
\textbf{CA-GRPO} & 55.4 & \textbf{68.1} & \textbf{55.2} & \textbf{67.3} \\
\bottomrule
\end{tabular}
\end{center}
\end{wraptable}

We compare CA-GRPO's confidence-adaptive reward mixing
(Equation~\ref{eq:adaptive-reward}) with three fixed outcome:process
ratios: $0.8{:}0.2$, $0.5{:}0.5$, and $0.2{:}0.8$.
Table~\ref{tab:reward-mixing-main} reports results on both benchmarks;
Appendix~\ref{app:reward-mixing-60step} provides the full comparison,
including evaluation with evolved decision knowledge.

The fixed mixtures exhibit a consistent trade-off between the two
decision levels. The process-heavy $0.2{:}0.8$ ratio yields the highest
fixed-mixture ELA on both benchmarks, whereas $0.5{:}0.5$ gives the
highest BDA. Moving from the latter to the former raises ELA by
1.4/0.8 percentage points on In-Period/Out-of-Period, but lowers BDA by
0.6/0.4 points. Thus, none of the tested fixed ratios is best for both
specific-label prediction and pass/violation decisions.

CA-GRPO achieves a better balance and ranks first in three of the four
comparisons. On In-Period, its ELA (55.4\%) is only 0.2 points below
the process-heavy mixture, while its BDA (\textbf{68.1\%}) is 0.8 points
higher. On Out-of-Period, it leads both ELA (\textbf{55.2\%}) and
BDA (\textbf{67.3\%}), exceeding the best fixed ratio for each metric
by 0.4 and 1.1 points, respectively. Adaptive mixing therefore retains
nearly all of the process-heavy mixture's In-Period ELA while improving
binary decisions, and yields the strongest observed generalization among
the evaluated mixtures.

This advantage persists with evolved decision knowledge
(Appendix~\ref{app:reward-mixing-60step}). CA-GRPO reaches 59.1\% ELA and
71.3\% BDA, exceeding the best fixed mixture for each metric by
0.5 and 1.2 points. Its BDA gain from knowledge is 3.2 points, compared
with 1.3--2.2 points for the fixed mixtures, indicating that adaptive
training remains beneficial when additional decision guidance is supplied.

\WFclear

\subsection{Knowledge and Rubric Updates}
\label{sec:analysis-experience-rubric}

\paragraph{Knowledge and continued training.}
Adding evolved knowledge to CA-GRPO improves In-Period performance
without updating the policy parameters (Table~\ref{tab:main-in-period}).
Continuing training with this knowledge and an updated process rubric
yields AdaptEvo, further improving ELA and BDA by 2.8 and 0.9 percentage
points, respectively. These gains reflect the combined effect of
continued training and rubric revision. We next examine why the process
evaluator itself may need to evolve during training.

\begin{figure}[t]
\centering
\includegraphics[width=\linewidth]{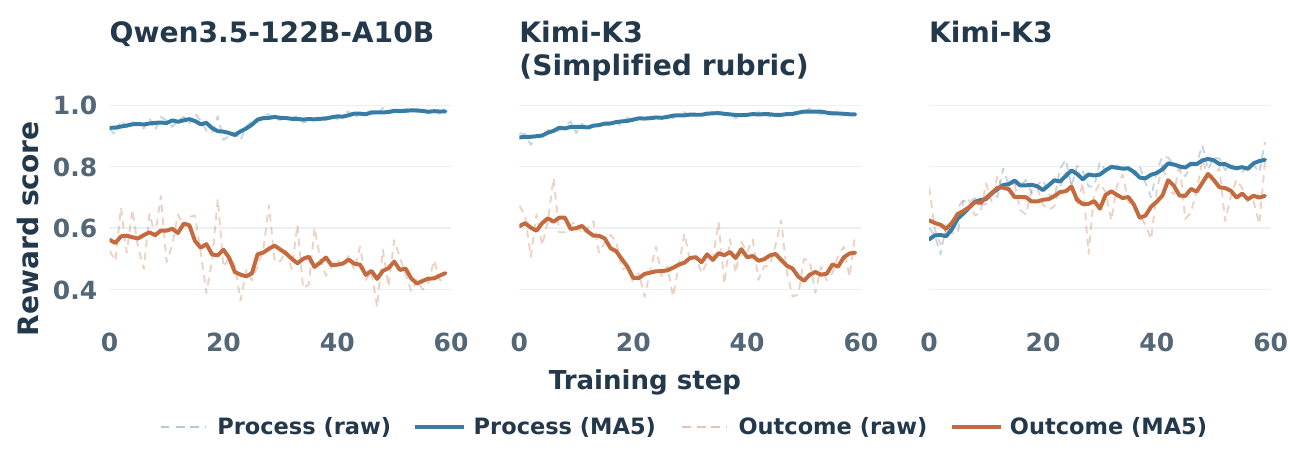}
\caption{Process (blue) and reference-agreement outcome (orange) rewards over
training steps 0--59. Panels identify process evaluators and mark the simplified
Kimi-K3 rubric. Faint dashed lines show raw step means; solid lines show
5-step centered moving averages (MA5) with truncated edge windows.
All panels share the same scale. Scores are unweighted training rewards,
not accuracies.}
\label{fig:process-outcome-diagnostics}
\end{figure}

\paragraph{Process-evaluator diagnostics.}
A process evaluator may fail to distinguish sound and defective rollouts, or a
permissive rubric may award high scores despite missing evidence checks
or incorrect rule application. Both can create exploitable scoring gaps,
allowing process reward to rise while task performance deteriorates
(Section~\ref{sec:process-reward-hacking}).
Figure~\ref{fig:process-outcome-diagnostics} shows a corresponding pattern:
with Qwen3.5-122B-A10B or Kimi-K3 using a simplified rubric, mean process
reward approaches 0.98, while reference-agreement outcome reward falls from
0.569 to 0.455 and from 0.615 to 0.464, respectively, between the first
and last 12-step windows. The other Kimi-K3 run improves on both rewards,
showing that evaluator identity alone does not determine the reward trends.

These observations motivate screening rubric checks for stable within-case
discrimination across multiple cases, accounting for repeated-judgment noise.
The divergence is consistent with process reward hacking, but differing
policy initializations and rubrics prevent isolating either failure
mechanism from these curves alone. Appendix~\ref{app:process-reward-diagnostics}
reports detailed statistics.

\section{Conclusion}
\label{sec:conclusion}
We presented AdaptEvo, a framework for learning under imperfect supervision
in rule-governed contextual decision tasks. AdaptEvo combines
confidence-adaptive policy optimization with cross-case knowledge synthesis
and rubric screening, addressing both uncertain
reference labels and gaps in rule application.
Experiments on industrial multimodal content moderation demonstrate
improved decision accuracy and gains from evolved decision knowledge.
Evaluation under changed rules further shows that CA-GRPO retains ELA
gains over the base model across evaluated checkpoints, whereas standard
GRPO deteriorates.

\subsection*{Contributors}
\label{sec:contributors}

\noindent
\mbox{Shijun Wan\textsuperscript{1,*}},
\mbox{Jiancong Xie\textsuperscript{2,*}},
\mbox{Hang Xu\textsuperscript{3}},
\mbox{Jin Duan\textsuperscript{3}},
\mbox{Qixiong Wang\textsuperscript{3}},
\mbox{Xi Xiang\textsuperscript{3}},
\mbox{Maofei Que\textsuperscript{3}},
\mbox{Yahui Liu\textsuperscript{3,\textdagger}},
\mbox{Zhongyu Wei\textsuperscript{1,\textdagger}},
\mbox{Mu Chuan\textsuperscript{3,\textdagger}}.

\begingroup
\small
\setlength{\parskip}{0pt}
\noindent
\textsuperscript{1}Fudan University\quad
\textsuperscript{2}Sun Yat-sen University\quad
\textsuperscript{3}Xiaohongshu\par
\smallskip
\noindent
\textsuperscript{*}Equal contribution.\quad
\textsuperscript{\textdagger}Corresponding authors.\par
\endgroup

\subsection*{AI use statement}
We use Qwen3.6-35B-A3B~\citep{qwen36_35b_a3b} as the shared policy
backbone. Our benchmark comparisons also include Qwen3.5-4B,
Qwen3.5-9B, and Qwen3.5-27B~\citep{qwen35_2026};
Qwen3.8-Flash-Next~\citep{qwen2026design};
GLM5.3-Flash~\citep{glm5team2026glm5};
Kimi-K3~\citep{kimi2026k3}; and
Gemini3.8-Flash~\citep{gemini2026flash38}.
Kimi-K3 also serves as a reference-support assessor and process evaluator;
the historical process-evaluator comparison additionally includes
Qwen3.5-122B-A10B~\citep{qwen35_2026}.
GLM5.3-Flash also scores candidate rubric checks during offline screening
and validation. GPT-5.6-Sol~\citep{openai2026gpt56sol} summarizes
the evolved decision knowledge. These research roles and their input
boundaries are specified in the method and experimental sections.

LLMs were also employed to assist in literature search, helping us identify
and locate relevant prior work. We also used
GPT-5.6-Sol~\citep{openai2026gpt56sol} to assist with implementing the
research code.

\bibliographystyle{styles/colm2026_conference}
\bibliography{bibliography/references}

\clearpage
\appendix
\section{Experimental Details}
\label{app:experimental-details}
\subsection{Moderation Task and Reference Supervision}
\paragraph{Content moderation instantiation.}
\label{sec:moderation-task}
Each moderation input $x=(c,s)$ consists of a post $c$ assigned to a queue
$s$. The post provides its title and body, available images or video key
frames, any OCR/ASR text, and metadata such as post type and author
information. Text and images are interleaved in the model input. The
queue determines the candidate-label set
$\mathcal{Y}(x)=\mathcal{L}_s=\{p_s\}\cup\mathcal{V}_s$, comprising one
pass label $p_s$ and violation labels $\mathcal{V}_s$.

The governing rules have two levels:
\begin{equation*}
 R=\bigl(R^{\mathrm{queue}}_s,
         \{(\ell,R_{s,\ell}):\ell\in\mathcal{V}_s\}\bigr).
\end{equation*}
Queue-wide rules $R^{\mathrm{queue}}_s$ specify the review scope and any
cross-label priorities; each $R_{s,\ell}$ provides the detailed rule for
violation label $\ell$, including its applicable conditions and any
exemptions. In the evaluated interface, the initial prompt contains the
full queue-wide rules and previews of the candidate violation rules.
These previews are deterministic extracts for navigation, not replacements
for the full rules. The queue-bound \texttt{get\_detail\_rule} tool retrieves
the full rules for requested violation labels. The pass label has no
separate violation rule; the prompt requires consulting at least one
relevant violation rule before returning pass.

Other tools in $\mathcal{T}$ retrieve image/video similarity results,
commercial details, author qualifications, recent posts, post comments,
profile-edit histories, reports, and the author's comments elsewhere.
Their schemas are supplied to the model, with the current case bound by
the runtime. Each request $a_k$ specifies a tool and its arguments; the
returned observation $o_k$ contains its status, text, and any images.
The agent finally
returns one label $\hat y\in\mathcal{L}_s$ and a rationale $j$, completing
the trajectory in Equation~\eqref{eq:problem-trajectory}.
Governing rules are distinct from optional evolved decision knowledge
$E_t$: omitting knowledge retains the rules and tool interface.
\paragraph{Example task instance.}
\label{app:moderation-task-instance}
Table~\ref{tab:moderation-task-instance} presents an anonymized instance
from an archived GRPO checkpoint evaluation without injected knowledge
($E_t=\varnothing$). It preserves the candidate-label mapping and recorded
tool-call order; post text, rule excerpts, observations, and the recorded
rationale are condensed and translated into English. The example
illustrates the task interface rather than a comparison between methods.

\begin{table}[!t]
\caption{A moderation task instance with two levels of governing rules
and a recorded tool-using decision. Label names and excerpts are translated;
identifying details are omitted.}
\label{tab:moderation-task-instance}
\begin{center}
\small
\renewcommand{\arraystretch}{1.12}
\begin{tabularx}{\linewidth}{@{}>{\raggedright\arraybackslash}p{0.19\linewidth}>{\raggedright\arraybackslash}X@{}}
\toprule
Component & Instance and correspondence \\
\midrule
Post $c$ and queue $s$ &
A mint-green eyeshadow post with product/swatch tags, nine post images,
and a transaction-linked commercial marker for Brand~A. Author metadata
also supplies an avatar and background image (11 image inputs in total).
The assigned queue reviews unfair competition in posts. \\
\addlinespace
Candidates $\mathcal{L}_s$ &
$\{p_s,v_1,v_2,v_3\}$, where $p_s=\text{Pass}$,
$v_1=\text{Unfair commercial comparison}$,
$v_2=\text{Commercial denigration}$, and
$v_3=\text{Personal negative review}$.
Here $v_1,v_2,v_3$ abbreviate this queue's violation labels. \\
\addlinespace
Queue rule $R_s^{\mathrm{queue}}$ &
Establish commercial context using account information and commercial
markers; assess the post and relevant author-originated context under
the applicable label rules. This queue-wide guidance is supplied in full
at initialization. \\
\addlinespace
Label rules $R_{s,\ell}$ &
For $v_1$, commercial applicability must be accompanied by covered unfair
comparison, such as unsupported scoring/ranking or one-sided competitor
criticism; balanced multi-product reviews can be exempt.
For $v_2$, assess competitive denigration and its prerequisites and
exemptions. For $v_3$, assess consumer negative claims, with exemptions
for specified subjective preferences or individual fit. Only previews
are initially shown; the full rules are retrieved below. \\
\midrule
Rule lookup $a_1\!\to\!o_1$ &
\texttt{get\_detail\_rule(labels=[}$v_1,v_2,v_3$\texttt{])}
returns the full rules for all three violation labels in the current
queue. The actual arguments are the corresponding label strings. \\
\addlinespace
Evidence lookup $a_2\!\to\!o_2$ &
\texttt{get\_commercial\_detail(\{\})} returns no associated product
items. The initial transaction-linked marker remains part of the evidence;
this empty result does not establish that the post is noncommercial. \\
\addlinespace
Recorded $(\hat y,j)$ &
$\hat y=p_s$. The recorded rationale describes a product and swatch
presentation without unfair rankings, denigration, or negative claims.
It therefore does not find the violation conditions of $v_1,v_2,v_3$
established; the commercial marker alone is insufficient.
The response cites both queried tools as evidence sources. \\
\bottomrule
\end{tabularx}
\end{center}
\end{table}

This correspondence separates \emph{which review criteria apply}
($R_s^{\mathrm{queue}}$) from \emph{what establishes a particular
violation} ($R_{s,\ell}$). The structured response stores $\hat y$ as
\texttt{predict\_label}, $j$ as \texttt{decision\_basis}, and optional
source identifiers as \texttt{used\_evidence\_ids}. The archived reference
label is also pass; reference labels and any human-review rationales are
offline information and are not supplied in the agent's inference prompt.

\paragraph{Human references and support assessment.}
\label{sec:moderation-supervision}
Each of the $3{,}395$ training examples has four human annotations,
including the quality-control (QC) reviewer's.
The QC reviewer's label is the final reference $y^\star$,
regardless of agreement among the four labels, and is accompanied by the
reviewer's written rationale. GT is an operational alias for this reference
in historical records and figures.

As the offline reference-support assessor, Kimi-K3 assesses the reference
label and rationale using the
complete post, applicable rules, and cached outputs from nine tool categories.
It checks whether evidence substantiates the stated rationale and whether
the assigned label follows from the rules, including necessary conditions
and exemptions. Violation decisions are checked against their label-specific
rules; pass decisions against all candidate violation rules in the queue.
For a violation reference, \texttt{supported} requires evidence for the
applicable rule and its necessary conditions, with no applicable exemption.
\texttt{Ambiguous} indicates substantive support but unresolved decisive
conditions, evidence, thresholds, or exemptions. \texttt{Unsupported}
indicates an inapplicable rule, a negated necessary condition, an applicable
exemption, or missing evidence explicitly required by the rule.
For a pass reference, \texttt{supported} means that no candidate violation
has sufficient support; \texttt{ambiguous} means that a plausible violation
remains unresolved; and \texttt{unsupported} requires at least one
established violation with no applicable exemption.
These assessments do not change the reference label. They measure support
for the recorded decision, without independently establishing a correct
label or estimating human annotation error.

\paragraph{Confidence and reward implementation.}
\label{sec:moderation-rewards}
For this instantiation, reference-label confidence combines human agreement
$C_{\mathrm{human}}(x)$ with assessed rule and evidence support for the QC label
and rationale, $G_{\mathrm{rule}}(x)$:
\begin{equation}
 C(x)=C_{\mathrm{human}}(x)G_{\mathrm{rule}}(x)\in[0,1].
 \label{eq:gt-confidence}
\end{equation}
Let $p_v$ be the fraction of the four annotations assigning normalized label
$v$. Human-agreement confidence is one minus normalized entropy,
\begin{equation}
 C_{\mathrm{human}}(x)=1-\frac{H(p)}{\log 4},
 \qquad H(p)=-\sum_v p_v\log p_v,
 \label{eq:human-confidence}
\end{equation}
with $0\log0=0$. Vote patterns $4$--$0$, $3$--$1$, $2$--$2$,
$2$--$1$--$1$, and $1$--$1$--$1$--$1$ map to $1$, approximately $0.5944$,
$0.5$, $0.25$, and $0$, respectively. Each missing annotation receives a
distinct unknown value, so missing rounds do not count as agreement.
The support gate $G_{\mathrm{rule}}$ is $1.0$, $0.6$, and $0.2$ for
\texttt{supported}, \texttt{ambiguous}, and \texttt{unsupported},
respectively; missing or unrecognized assessments use $0.6$.
Neither factor is assumed to be a calibrated correctness probability.
The reward mixture uses $\alpha(C)=0.9C$ throughout the reported method.

For a valid single-label prediction, the outcome reward is
\begin{equation}
 R_{\mathrm{out}}(\hat y,y^\star)=
 \begin{cases}
  1, & \hat y=y^\star,\\
  0.6, & \hat y\ne y^\star\text{ and }\hat y,y^\star\in\mathcal V_s,\\
  0, & \text{otherwise}.
 \end{cases}
 \label{eq:moderation-outcome-reward}
\end{equation}
Comparison normalizes label strings by removing an optional numeric
identifier prefix. Thus a wrong violation subtype receives partial
reward, whereas a pass/violation error does not. This graded training
reward is distinct from both ELA and BDA.

The agent's structured-output schema requests one candidate label and a
nonempty rationale. The final-output parser rejects malformed or empty
structured outputs, missing labels, and labels outside $\mathcal L_s$;
a list of multiple labels is invalid. An invalid final output overrides
the mixed reward to zero. For valid outputs, let $k(\tau)$ count unknown
tool calls, malformed tool names, invalid tool arguments, and other
recorded interaction-format errors. The protocol penalty is
$P_{\mathrm{protocol}}(\tau)=\min(0.3,0.1k(\tau))$ and is subtracted as
in Equation~\eqref{eq:adaptive-reward}. The implementation also applies
a total-reward floor of $-1$, which is inactive under these reward and
penalty bounds.

The process reward in Equation~\eqref{eq:process-reward} uses five
dimensions: factual grounding, rule fidelity, evidence coverage, tool use,
and decision consistency, with weights $(0.25,0.25,0.20,0.15,0.15)$.
The process evaluator emits one score per dimension directly on $[0,1]$;
the parser rejects missing fields, nonnumeric or nonfinite values, and
out-of-range scores before computing the weighted sum. Qualitative
checks and scoring anchors guide each dimension score, with no numerical
check-level aggregation (Appendix~\ref{app:rubric-schema}).
The evaluator scores observable evidence and trajectories without the raw
reference answer by default. This role differs from reference-support
assessment even when the underlying model is the same.

If a required process score remains unavailable after retries for any
otherwise valid trajectory, the training filter rejects its entire
rollout group. Invalid final outputs remain trainable at zero reward and
do not require a process score. Missing process scores therefore are not
treated as observed zero-quality trajectories.

\subsection{Human Agreement and Reference Support}
\label{app:supervision-diagnostics}
The annotation and support-assessment protocol is defined in
Appendix~\ref{sec:moderation-supervision}. Table~\ref{tab:human-agreement-gt-support}
cross-tabulates exact-label agreement among the four human annotations and
Kimi-K3's support assessment of the QC label and rationale.

\begin{table}[htbp]
\renewcommand{\arraystretch}{1.12}
\caption{Human agreement and assessed support for QC decisions on the training set.}
\label{tab:human-agreement-gt-support}
\begin{center}
\small
\begin{tabular*}{\linewidth}{@{\extracolsep{\fill}}lrrrrr@{}}
\toprule
Agr. & S & A & U & $n$ & \% \\
\midrule
\texttt{4-0} & \textbf{1,968} & 473 & 245 & 2,686 & 79.12 \\
\texttt{3-1} & 287 & 175 & 212 & 674 & 19.85 \\
\texttt{2-2} & 11 & 7 & 6 & 24 & 0.71 \\
\texttt{2-1-1} & 5 & 4 & 1 & 10 & 0.29 \\
\texttt{1-1-1-1} & 0 & 0 & 1 & 1 & 0.03 \\
\midrule
\textbf{Total} & \textbf{2,271} & \textbf{659} & \textbf{465} & \textbf{3,395} & \textbf{100.00} \\
\bottomrule
\end{tabular*}
\par\smallskip
\begin{minipage}{\linewidth}\footnotesize
Agr.: agreement pattern across four human labels; S/A/U: supported, ambiguous,
or unsupported; $n$: row total; \%: share of $N=3{,}395$ samples.
The QC label remains the reference under every agreement pattern.
\end{minipage}
\end{center}
\end{table}

Overall, $709$ of $3{,}395$ cases (20.9\%) have nonunanimous human labels.
Kimi-K3 rates $465$ cases (13.7\%) as unsupported and $659$ (19.4\%) as
ambiguous, totaling $1{,}124$ (33.1\%) in these two support categories.
All four percentages use the full training set as the denominator.
The disagreement and ambiguous/unsupported groups overlap, so their
percentages should not be added.

Of $2{,}686$ unanimous cases, $718$ (26.73\%) are rated ambiguous or
unsupported; $303$ of $709$ nonunanimous cases are rated supported.
The intersection of unanimity and support contains $1{,}968$ samples
(57.97\% of the training set), which defines the high-confidence subset.
These statistics show that agreement and assessed support are distinct
signals in this dataset. They establish neither the annotation error rate
nor a relationship between confidence and task difficulty.

\subsection{Training and Evaluation Protocol}
\paragraph{Models and comparison methods.}
\label{sec:experiments-baselines}

Table~\ref{tab:planned-variants} summarizes the shared-backbone settings
defined in Section~\ref{sec:experiments-setup}; internal run identifiers
appear in Appendix~\ref{app:reproducibility}. The full training pool contains
$3{,}395$ examples, and the high-confidence subset contains $1{,}968$
(Appendix~\ref{app:supervision-diagnostics}).

\begin{table}[H]
\renewcommand{\arraystretch}{1.12}
\caption{Training settings with a shared Qwen3.6-35B-A3B initialization. AdaptEvo continues from CA-GRPO after a joint knowledge and rubric update.}
\label{tab:planned-variants}
\begin{center}
\small
\begin{tabularx}{\linewidth}{@{}p{0.30\linewidth}p{0.17\linewidth}X@{}}
\toprule
Setting & Training pool & Supervision and evolution \\
\midrule
Base Model & --- & No task-specific RL \\
GRPO & Full pool & Reference-based outcome rewards \\
GRPO (High-Conf.) & High-confidence subset & Reference-based outcome rewards \\
CA-GRPO & Full pool & Confidence-adaptive outcome/process mixture; no injected decision knowledge; fixed rubric \\
AdaptEvo & Full pool & Continued CA-GRPO after joint knowledge/rubric evolution \\
\bottomrule
\end{tabularx}
\end{center}
\end{table}

The knowledge-only row in the main table reuses the CA-GRPO checkpoint,
adding evolved knowledge at inference without further parameter updates.
The recorded knowledge summarizer is \textbf{GPT-5.6-Sol}, distinct from
the policy backbone. The AdaptEvo results cover one joint knowledge--rubric
update followed by continued training.

\paragraph{Data availability.}
We will publicly release the training and evaluation datasets.

\paragraph{Evaluation datasets.}
The main table reports results on the In-Period and Out-of-Period
benchmarks defined in Section~\ref{sec:experiments-setup}.
Each of the 13 model/knowledge configurations has ELA and BDA results
on both benchmarks. Table~\ref{tab:shared-backbone-paired} consolidates
the six shared-backbone configurations and the three fixed reward
mixtures, with knowledge injection specified separately. The fixed
ratios are outcome:process weights; Appendix~\ref{app:reward-mixing-60step}
also reports their In-Period results with knowledge at inference.
Section~\ref{sec:analysis-generalization}
evaluates the base model and six checkpoints per method on the same
Out-of-Period benchmark, at completed training steps 10--60 with an interval
of 10 steps. These checkpoint evaluations omit injected decision knowledge;
their step-60 results match the corresponding main-table rows.

\begin{table}[htbp]
\setlength{\tabcolsep}{4pt}
\renewcommand{\arraystretch}{1.10}
\caption{Shared-backbone results on In-Period and Out-of-Period
(\%, higher is better). All configurations use Qwen3.6-35B-A3B.
Fixed ratios specify outcome:process reward weights; Knowledge indicates
injected decision knowledge. Bold marks the best score in each column.}
\label{tab:shared-backbone-paired}
\begin{center}
\small
\begin{tabularx}{\linewidth}{@{}lc*{4}{>{\centering\arraybackslash}X}@{}}
\toprule
\textbf{Method} & \textbf{Knowledge}
& \multicolumn{2}{c}{\textbf{In-Period}}
& \multicolumn{2}{c@{}}{\textbf{Out-of-Period}} \\
\cmidrule(lr){3-4}\cmidrule(l){5-6}
& & \textbf{ELA} & \textbf{BDA} & \textbf{ELA} & \textbf{BDA} \\
\midrule
Base model & None & 53.9 & 66.1 & 53.3 & 65.9 \\
GRPO & None & 54.4 & 68.5 & 51.9 & 65.2 \\
GRPO (High-Conf.) & None & 53.3 & 68.8 & 52.9 & 65.5 \\
\midrule
Fixed (0.8:0.2) & None & 54.3 & 67.6 & 53.6 & 66.1 \\
Fixed (0.5:0.5) & None & 54.2 & 67.9 & 54.0 & 66.2 \\
Fixed (0.2:0.8) & None & 55.6 & 67.3 & 54.8 & 65.8 \\
CA-GRPO & None & 55.4 & 68.1 & 55.2 & 67.3 \\
\midrule
CA-GRPO$^{\dagger}$ & Evolved & 59.1 & 71.3 & 58.6 & \textbf{69.7} \\
\rowcolor{starlight!45}[0pt][0pt]
\textbf{AdaptEvo} & Evolved & \textbf{61.9} & \textbf{72.2} & \textbf{59.5} & 69.5 \\
\bottomrule
\end{tabularx}
\par\smallskip
\begin{minipage}{\linewidth}
\footnotesize
Trained methods use their reported 60-step checkpoints.
$^{\dagger}$Knowledge is added only at inference to the same CA-GRPO
checkpoint. AdaptEvo continues training with evolved knowledge and an
updated process rubric. Values agree with
Tables~\ref{tab:main-in-period} and~\ref{tab:reward-mixing-60step}.
\end{minipage}
\end{center}
\end{table}

\paragraph{Training budget.}
\label{app:training-budget}
In the logged training configuration, each step retains 32 prompt groups,
with eight rollout trajectories per group. A retained prompt occurrence is
one such group: the same post can contribute multiple occurrences across
steps. Each 10-step checkpoint interval therefore comprises 320 retained
prompt occurrences and $2{,}560$ retained trajectories, not 320 unique posts.
Rejected or retried generations can incur additional generation costs.
The total AdaptEvo budget includes the initial CA-GRPO training
as well as training after Evolution. Filtering the training pool changes both sample
composition and exposure, so equal steps alone do not establish equal data
coverage.

\paragraph{Checkpoint convention and comparison conditions.}
The main table reports the 60-step checkpoint for each trained method;
the knowledge-only comparison reuses that CA-GRPO checkpoint. This convention
uses run-local steps; AdaptEvo includes initial and continued training. Each table row uses one checkpoint for
all metrics. The continuation comparison combines additional policy
training with rubric revision and does not isolate their individual effects.

\subsection{Metric Definitions}
\begingroup
\displaywidowpenalty=10000
\paragraph{Evaluation metrics.}
\label{sec:experiments-metrics}

We use exact-label accuracy (ELA) and binary decision accuracy (BDA) across
the evaluation sets.
For $N$ evaluation samples, let $s_i$ be the queue of sample $i$, with reference and model labels
$y_i^\star,\hat y_i\in\mathcal{L}_{s_i}$.
Define~$b_s(\ell)=\mathbb{I}[\ell\in\mathcal{V}_s]$, mapping the queue's pass label to $0$ and every violation label to $1$.
The two accuracies are
\begin{align}
\mathrm{ELA} &= \frac{1}{N}\sum_{i=1}^{N}\mathbb{I}[\hat y_i=y_i^\star], \\
\mathrm{BDA} &= \frac{1}{N}\sum_{i=1}^{N}\mathbb{I}[b_{s_i}(\hat y_i)=b_{s_i}(y_i^\star)].
\end{align}
\endgroup
For example, if the reference is
one violation label $v_{s,1}$ and the prediction is a different violation
label $v_{s,2}$ in the same queue, the example counts as incorrect for ELA
but correct for BDA. Predicting the pass label $p_s$ for that reference
counts as incorrect for both metrics. An exact label match is correct for
both. Thus, BDA measures the coarser moderation decision, while ELA also
requires the correct violation category.

The formulas define $0$--$1$ proportions; tables multiply these by $100$ to display percentages, with changes expressed in \textbf{percentage points}.
Training reward scores in Figure~\ref{fig:process-outcome-diagnostics} retain their original units.

Current single-run summaries establish neither stability nor statistical
significance.
Process-quality comparisons require a fixed evaluation reference; raw rubric
scores do not share a common scale across versions.

\subsection{Evaluation with New Rules and Queues}
\label{app:shifted-evaluation}

Table~\ref{tab:shifted-evaluation} reports the Out-of-Period checkpoint
evaluation on the same benchmark as Table~\ref{tab:main-in-period}.
Its base and 60-step results agree with the corresponding main-table rows.
All agents run
without injected decision knowledge, and no stronger model synthesizes
knowledge for this evaluation. The run aliases
\texttt{gtdirect-full}, \texttt{gtdirect-support}, and \texttt{gtconf-v1}
correspond to GRPO, GRPO (High-Conf.), and CA-GRPO, respectively.
We report completed training steps: checkpoint indices
9, 19, 29, 39, 49, and 59 correspond to steps 10, 20, 30, 40, 50, and 60.
The source records retain these original checkpoint indices.
Figure~\ref{fig:shifted-generalization}(b--c) displays each checkpoint's
accuracy minus the corresponding base accuracy (53.3\% ELA or 65.9\% BDA),
in percentage points. Both heatmaps use the same color scale centered
at zero; the checkpoint values are not smoothed.

\begin{table}[htbp]
\renewcommand{\arraystretch}{1.12}
\caption{Out-of-Period checkpoint results without injected decision
knowledge. All methods initialize from the same
Qwen3.6-35B-A3B base model. Accuracies are percentages at the supplied
one-decimal precision; bold marks the highest checkpoint result per metric.}
\label{tab:shifted-evaluation}
\begin{center}
\small
\begin{tabularx}{\linewidth}{@{}l*{6}{>{\centering\arraybackslash}X}@{}}
\toprule
& \multicolumn{2}{c}{GRPO} & \multicolumn{2}{c}{GRPO (High-Conf.)}
& \multicolumn{2}{c}{CA-GRPO} \\
\cmidrule(lr){2-3}\cmidrule(lr){4-5}\cmidrule(l){6-7}
Training steps & ELA & BDA & ELA & BDA & ELA & BDA \\
\midrule
Base & 53.3 & 65.9 & 53.3 & 65.9 & 53.3 & 65.9 \\
\midrule
10 & 54.5 & 66.0 & 55.1 & 66.8 & 54.1 & 66.2 \\
20 & 54.2 & 66.2 & 54.4 & 67.2 & 54.3 & 65.7 \\
30 & 54.2 & 66.5 & 53.9 & 66.8 & 54.5 & 65.9 \\
40 & 53.3 & 65.9 & 52.0 & 65.1 & 54.4 & 65.7 \\
50 & 52.2 & 65.5 & 52.5 & 65.3 & 54.8 & 66.2 \\
60 & 51.9 & 65.2 & 52.9 & 65.5 & \textbf{55.2} & \textbf{67.3} \\
\bottomrule
\end{tabularx}
\end{center}
\end{table}

\paragraph{Training-reward processing.}
Figure~\ref{fig:shifted-generalization}(a) uses the recorded outcome
rewards at steps 10--60: \texttt{full} denotes GRPO and
\texttt{supported} denotes GRPO (High-Conf.). These training-log steps
are plotted as recorded, without the checkpoint-index offset above.
The solid curves are centered 5-step moving averages computed from the
raw rewards over the displayed interval, steps 10--60. Boundary windows
use only the available observations within that interval, following the
same smoothing convention as the process-evaluator diagnostic.
Separately, mean raw rewards over the 10-step windows 10--19 and
51--60 increase from 0.6135 to 0.6238 for GRPO and from 0.7109 to 0.7550
for GRPO (High-Conf.), although neither trajectory is monotonic.
The two methods train on different data subsets, so their absolute
reward levels are not a comparison on the same examples.

The reported results summarize aggregate ELA and BDA; they do not establish
per-queue effects, variability across independent runs, or uncertainty
at the sample level.
Changed rules and new queues are evaluated jointly; the result does not
separate their effects.

\subsection{Reward-Mixture Comparison at 60 Training Steps}
\label{app:reward-mixing-60step}

Table~\ref{tab:reward-mixing-60step} reports the supplied fixed-ratio
comparison. The source aliases $0.8/0.2$, $0.5/0.5$, and $0.2/0.8$
denote fixed outcome:process reward weights; \texttt{adapt} denotes
CA-GRPO. Every result is measured after 60 training steps on the same
In-Period and Out-of-Period benchmarks used in the main table. The knowledge
condition adds decision knowledge at inference on In-Period. The CA-GRPO
row uses the same 60-step checkpoint as the main table and
Appendix~\ref{app:shifted-evaluation}.

\begin{table}[htbp]
\setlength{\tabcolsep}{5pt}
\renewcommand{\arraystretch}{1.12}
\caption{Fixed and adaptive outcome/process reward mixing after 60 training steps.
All values are percentages. Bold marks the best value in each column.}
\label{tab:reward-mixing-60step}
\begin{center}
\small
\begin{tabularx}{\linewidth}{@{}Xrrrrrr@{}}
\toprule
& \multicolumn{2}{c}{In-Period} & \multicolumn{2}{c}{In-Period + knowledge} & \multicolumn{2}{c}{Out-of-Period} \\
\cmidrule(lr){2-3}\cmidrule(lr){4-5}\cmidrule(l){6-7}
Outcome : process & ELA & BDA & ELA & BDA & ELA & BDA \\
\midrule
0.8:0.2 & 54.3 & 67.6 & 57.7 & 68.9 & 53.6 & 66.1 \\
0.5:0.5 & 54.2 & 67.9 & 58.1 & 70.1 & 54.0 & 66.2 \\
0.2:0.8 & \textbf{55.6} & 67.3 & 58.6 & 69.3 & 54.8 & 65.8 \\
\rowcolor{starlight!45}[0pt][0pt]
CA-GRPO (adaptive) & 55.4 & \textbf{68.1} & \textbf{59.1} & \textbf{71.3} & \textbf{55.2} & \textbf{67.3} \\
\bottomrule
\end{tabularx}
\end{center}
\end{table}

Comparisons to the best fixed ratio take the maximum separately for each
metric and condition. The $0.2{:}0.8$ ratio has the highest fixed-mixture
ELA in all three conditions, whereas $0.5{:}0.5$ has the highest
fixed-mixture BDA. These two maxima do not represent a single configuration.
All values are reported to one decimal place.

Equal training steps are confirmed for this comparison, but equal steps
alone do not establish equal data exposure or total compute.

\subsection{Run Aliases and Checkpoint Conventions}
\label{app:reproducibility}
Table~\ref{tab:run-name-map} records the internal run aliases, while the main table
uses compact model/method names and a separate Knowledge column.
The six shared-backbone configurations use Qwen3.6-35B-A3B as the
policy base; the recorded knowledge summarizer is GPT-5.6-Sol.
Model names in Figure~\ref{fig:process-outcome-diagnostics} identify process evaluators.
Those three historical diagnostic runs do not all share the same policy
initialization and are separate from the six main configurations.

\begin{table}[htbp]
\caption{Mapping from internal run aliases to the compact main-table
names and knowledge settings.}
\label{tab:run-name-map}
\begin{center}
\small
\begin{tabularx}{\linewidth}{@{}lXl@{}}
\toprule
Internal alias & Main-table name & Knowledge \\
\midrule
\texttt{base} & Qwen3.6-35B-A3B (base) & None \\
\texttt{gtdirect-full} & GRPO & None \\
\texttt{gtdirect-support} & GRPO (High-Conf.) & None \\
\texttt{gtconf-v1}, library off & CA-GRPO & None \\
\texttt{gtconf-v1}, library on & CA-GRPO & Evolved \\
\texttt{gtconf-v2} & AdaptEvo & Evolved \\
\bottomrule
\end{tabularx}
\end{center}
\end{table}

All 13 model/knowledge configurations in the main table have results on
both benchmarks. The seven reference models from the Qwen, GLM, Kimi, and
Gemini families are evaluated without injected decision knowledge.
Model names alone do not establish exact serving versions or task-training
settings.

The main table reports each trained method's 60-step checkpoint, with
percentages to one decimal place. The knowledge-only row reuses the same
CA-GRPO checkpoint. Table~\ref{tab:shifted-evaluation} follows six checkpoints
at completed training steps 10--60 on the same Out-of-Period benchmark;
all agents in that comparison run without injected knowledge.
Table~\ref{tab:reward-mixing-60step} also reports a 60-step reward-mixture
comparison. These are run-local steps; AdaptEvo includes initial and
continued training.
Prompt and trajectory counts follow the retained-group convention in
Appendix~\ref{app:training-budget}.

\section{Method Protocol and Artifact Schemas}
\label{app:artifacts}
\subsection{Method Protocol and Diagnostic Details}
\label{app:method-details}

\paragraph{Stage-transition equations.}
\label{sec:alternating-optimization}

At predefined stage boundaries, AdaptEvo alternates policy learning,
failure diagnosis and grouping, and knowledge--rubric selection:
\begin{equation}
    \begin{aligned}
        \theta_{t+1}
            &= \mathcal{U}_{\mathrm{RL}}(\theta_t;E_t,Q_t,C), \\
        (F_t,\mathcal{G}_t)
            &= \mathcal{D}(\theta_{t+1};E_t,Q_t,R,\mathcal{T}), \\
        (E_{t+1},Q_{t+1})
            &= \mathcal{S}\bigl(E_t,Q_t,
                 \mathcal{U}_E(E_t,\mathcal{G}_t;M_{\mathrm{strong}}),\mathcal{U}_Q(Q_t,\mathcal{B}_t;\mathcal{P})\bigr).
    \end{aligned}
    \label{eq:alternating-update}
\end{equation}
Here $\mathcal{U}_{\mathrm{RL}}$ updates the policy, $\mathcal{D}$ diagnoses
and groups failures, and $\mathcal{U}_E$ and $\mathcal{U}_Q$ are the proposal
operators in Section~\ref{sec:method-evolution}. Rubric screening uses
within-input rollout groups $\mathcal{B}_t$ and candidate criteria
$\mathcal{P}$, separately from the cross-case failure groups $\mathcal{G}_t$.
The selector $\mathcal{S}$
chooses knowledge and rubric versions and may retain either previous version. The external
updater and evaluator need not be trained in this loop. Each accepted
update changes the guidance or feedback for the next policy stage, whose
remaining failures inform subsequent revisions. Our experiments cover two
training stages and one joint knowledge--rubric update. The following paragraphs describe queue-level synthesis and the update interfaces.

\paragraph{Queue-level knowledge extraction.}
The workflow processes training cases from one moderation queue jointly.
Error-pattern groups organize cases within the queue; labels may be
compared to clarify boundaries but do not define independent optimization
tasks. Cross-case recurrence identifies patterns shared by different
inputs. Attribution separates decision defects from reference conflicts,
missing information, and tool or cache failures. Repeated disagreement
with a reference is not by itself sufficient grounds for a knowledge
entry. Guidance is stored and deployed per queue; injected text contains
reusable conditions rather than case identifiers or individual answers.

\paragraph{Cross-case knowledge synthesis.}
The synthesis workflow collects training cases with the same diagnosed
failure mechanism, identifies their shared evidence patterns and missed
conditions, and contrasts cases that differ in context, evidence
sufficiency, exemptions, or adjacent-label boundaries. Human rationales
help identify these differences but do not authorize new rules or hidden
thresholds. A distinction enters guidance only when the governing criteria
and observable evidence support it; unexplained reference differences are
recorded as conflicts or subjective boundaries. The model then writes
conditional guidance with an explicit scope and consolidates compatible
findings across the queue. This comparison is intended to avoid turning
individual reviewer preferences into unconditional instructions.

\paragraph{Knowledge and rubric updates.}
Knowledge entries and supporting records follow the schema in
Appendix~\ref{app:experience-schema}. Updates may add conditions, refine
exceptions, consolidate overlapping guidance, or remove invalid entries.
The method distinguishes two assessment interfaces: knowledge changes
require fresh agent rollouts because they can alter evidence acquisition
and tool use, whereas rubric changes can be assessed on fixed trajectories
with the evaluator held fixed. Increased agreement between new guidance
and a new rubric alone does not establish better rule application.
The rubric screening and selected changes appear in
Appendix~\ref{app:rubric-schema}.

\paragraph{Update scope.}
Accepted knowledge and rubric versions remain fixed throughout the next
training stage. Governing rules are preserved, and final test cases and
outcomes are excluded from extraction, diagnosis, and update selection.
Cross-case synthesis targets reusable decision conditions but does not by
itself guarantee transfer. The evaluated run aliases and knowledge
settings are recorded in Appendix~\ref{app:reproducibility}.

\subsection{Decision Knowledge Entry Structure}
\label{app:experience-schema}
The knowledge library is stored and deployed per moderation queue.
Each entry records its scope, triggering conditions, required evidence,
decision conditions, exceptions, and applicability limits. These fields
specify reusable guidance for the agent. Supporting records attach
provenance, case identifiers, original evidence, and version information;
these records remain outside the injected text. Version records retain
the accepted content and the reason for accepting or rejecting a candidate.
Supporting records also preserve unresolved rule--reference conflicts and
subjective boundaries, which are not promoted into new decision standards.
The extraction procedure appears in
Appendix~\ref{app:method-details}.

\subsection{Rubric Checks and Scoring Anchors}
\label{app:rubric-schema}
\label{app:rubric-screening}
Each rubric version records qualitative checks, applicability conditions,
and scoring anchors within the fixed process dimensions in
Appendix~\ref{sec:moderation-rewards}. Checks specify what evidence and
behavior to inspect; anchors guide the resulting dimension score.
During policy training, the evaluator returns five dimension scores, with no numerical
subscores for individual checks. The reward code computes their weighted
sum. Evolution changes these scoring instructions while preserving the
dimension set and weights.

\paragraph{Screening protocol.}
GLM5.3-Flash scored 19 candidate checks in batches on four saved rollouts
for each of 140 training cases across seven queues, without reference
labels, confidence scores, or human rationales. Screening compared
within-case rollout variance across cases with repeated-judgment noise,
alongside mean scores and full-score rates. The selected checks were
re-evaluated on another 140 training cases, with two rollouts per case
and three judgments per rollout, before integration into the rubric.

\paragraph{Versioned check revisions.}
The revised rubric strengthens three named checks.
\textbf{R01} requires verifying decisive rule conditions, including
conjunctions, disjunctions, evidence attribution, and exceptions; naming a
label or retrieving a rule does not establish that its conditions were met.
\textbf{T03} checks whether the decision actually uses returned tool evidence
and distinguishes no result, explicit negative evidence, request failure,
cache unavailability, and insufficient evidence.
\textbf{T04} checks for verifiably redundant or uninformative tool calls,
while allowing reasonable retries and calls that resolve new questions.
It does not penalize call count, token length, or elapsed time by itself.
T03 and T04 inform the single tool-use score; they produce neither separate
output fields nor additive subitem penalties.

\paragraph{Scoring anchors.}
The initial rubric provides example anchors at $0$, $0.2$, $0.5$, $0.8$,
and $1$ with specified score intervals. The revised rubric instructs the
evaluator to use six anchored levels: $1$ for reliable behavior with no
substantive defect; $0.8$ for a minor local issue that preserves the core
evidence chain; $0.6$ for an important gap with substantial valid evidence;
$0.4$ for multiple important defects and only partial support; $0.2$ for
a severe verifiable error invalidating most of the dimension; and $0$ for
verifiable near-complete failure. Missing or unobservable information is
not automatically a zero score. The common parser validates finite scores
in $[0,1]$, but does not enforce membership in the six-level set; the
anchors are evaluator instructions, not a deterministic quantization step.
Unavailable evaluator scores are handled by the rollout-group filter in
Appendix~\ref{sec:moderation-rewards}.

\clearpage
\section{Additional Moderation Diagnostics}
\label{app:moderation-diagnostics}
These diagnostics complement the main evaluation and do not isolate the
effects of individual framework components.
\subsection{Historical Process--Outcome Reward Diagnostics}
\label{app:process-reward-diagnostics}

Figure~\ref{fig:process-outcome-diagnostics} compares three runs over steps 0--59;
panel names identify the process evaluators and mark the simplified Kimi-K3 rubric.
Between the first and last 12-step windows (steps 0--11 and 48--59),
mean process reward in the run scored by Qwen3.5-122B-A10B rises from 0.936 to 0.981
while mean reference-agreement outcome reward falls from 0.569 to 0.455.
For Kimi-K3 with a simplified process rubric, process reward increases from 0.914 to 0.975
while reference-agreement outcome reward falls from 0.615 to 0.464.
Both runs approach the process-reward ceiling while outcome rewards decline.

By contrast, the other Kimi-K3 run improves on both rewards over the same
interval: process reward rises from 0.635 to 0.813, while reference-agreement outcome reward
rises from 0.652 to 0.731. The two Kimi-K3 runs exhibit different reward
trends despite using the same evaluator model, motivating examination of the
rubric and training configuration rather than the evaluator identity alone.

The Qwen3.5-122B-A10B and simplified-rubric Kimi-K3 patterns motivate the concern in
Section~\ref{sec:process-reward-hacking}: limited evaluator capability may
prevent reliable discrimination between rollouts, while an overly
permissive rubric may give high scores despite critical errors.
Either failure can create exploitable scoring gaps, allowing optimization
to increase process reward while outcomes regress.
Here, permissiveness means failing to distinguish defective trajectories
from sound ones; a high mean score alone does not establish this failure.

These are unweighted raw training rewards, not accuracies. The reference-agreement
outcome reward is also an imperfect proxy, rather than an independent
assessment of true decision quality. The runs differ in policy
initialization and rubric configuration. Consequently,
their aggregate curves neither isolate evaluator permissiveness as the
cause nor establish a causal ranking of evaluators. The historical curves
alone do not establish that rubric updates prevent reward hacking.

\end{document}